\documentclass{article}

\usepackage{microtype}
\usepackage{graphicx}
\usepackage{subfigure}
\usepackage{booktabs} % for professional tables

\usepackage{hyperref}

\usepackage[accepted]{icml2025}

\usepackage{amsmath}
\usepackage{amssymb}
\usepackage{mathtools}
\usepackage{amsthm}

\usepackage[capitalize,noabbrev]{cleveref}

\theoremstyle{plain}

\theoremstyle{definition}

\theoremstyle{remark}

\usepackage[textsize=tiny]{todonotes}

\icmltitlerunning{Submission and Formatting Instructions for ICML 2025}

\begin{document}

\twocolumn[
\icmltitle{Beyond Nearest Neighbors: Semantic Compression and Graph-Augmented Retrieval for Enhanced Vector Search}




\begin{icmlauthorlist}
\icmlauthor{Rahul Raja}{cmu,linkedin,stanford}
\icmlauthor{Arpita Vats}{bu,linkedin}
\end{icmlauthorlist}

\icmlaffiliation{linkedin}{LinkedIn, California, USA}
\icmlaffiliation{bu}{Boston University, Boston, USA}
\icmlaffiliation{cmu}{Carnegie Mellon University, Pittsburgh, USA}
\icmlaffiliation{stanford}{Stanford University, Palo Alto, USA}

\icmlcorrespondingauthor{Rahul Raja}{rauhl.110392@gmail.com}
\icmlcorrespondingauthor{Arpita Vats}{arpita.vats09@gmail.com}
\icmlkeywords{Machine Learning, ICML}

\vskip 0.3in
]



\printAffiliationsAndNotice{Work does not relate to position at LinkedIn.}
% \footnotetext[5]{\url{https://github.com/rahulrj/icml_vecdb_experiments}}
% \footnotetext[1]{\url{https://github.com/rahulrj/icml_vecdb_experiments}}

\begin{abstract}
Vector databases typically rely on approximate nearest neighbor (ANN) search to retrieve the top-\(k\) closest vectors to a query in embedding space. While effective, this approach often yields semantically redundant results, missing the diversity and contextual richness required by applications such as retrieval-augmented generation (RAG), multi-hop QA, and memory-augmented agents. We introduce a new retrieval paradigm: \textit{semantic compression}, which aims to select a compact, representative set of vectors that captures the broader semantic structure around a query. We formalize this objective using principles from submodular optimization and information geometry, and show that it generalizes traditional top-\(k\) retrieval by prioritizing coverage and diversity. To operationalize this idea, we propose \textit{graph-augmented vector retrieval}, which overlays semantic graphs (e.g., kNN or knowledge-based links) atop vector spaces to enable multi-hop, context-aware search. We theoretically analyze the limitations of proximity-based retrieval under high-dimensional concentration and highlight how graph structures can improve semantic coverage. Our work outlines a foundation for meaning-centric vector search systems, emphasizing hybrid indexing, diversity-aware querying, and structured semantic retrieval. We make our implementation publicly available to foster future research in this area\textsuperscript{5}.
\footnotetext[5]{\url{https://github.com/rahulrj/icml_vecdb_experiments}}
\end{abstract}

\section{Introduction}

Vector search systems, often powered by approximate nearest neighbor (ANN) algorithms~\citep{johnson2019billion}, are widely used for tasks such as semantic search~\citep{guo2020accelerating}, recommendation~\citep{wang2021survey}, and retrieval-augmented generation (RAG)~\citep{lewis2020retrieval}. These systems typically retrieve the top-$k$ vectors closest to a query in embedding space. While effective for ensuring local relevance, this approach frequently results in semantically redundant outputs that lack contextual diversity—an important limitation for multi-hop QA, fact-based summarization, and memory-augmented reasoning~\citep{asai2020learning}.

We propose a new retrieval paradigm: \textit{semantic compression}, which aims to return a compact and representative set of vectors that captures the broader semantic structure around a query. Unlike conventional top-$k$ retrieval that prioritizes proximity, semantic compression emphasizes both relevance and diversity by framing retrieval as a set selection problem. We formalize this objective using submodular optimization~\citep{krause2014submodular} and information geometry, showing it to be a generalization of proximity-based methods.

To make semantic compression practical, we introduce \textbf{graph-augmented vector retrieval}, where symbolic edges (e.g., kNN, clustering, or knowledge-based links) are added to embedding spaces~\citep{liu2021learning}. These graphs enable context-aware, multi-hop retrieval methods such as personalized PageRank (PPR)~\citep{haveliwala2003topic}, allowing the discovery of semantically diverse but non-local results. Our evaluations demonstrate that graph-based methods, especially with dense symbolic connections, outperform pure ANN retrieval in semantic diversity while maintaining high relevance.

The main contributions of this paper are:
\begin{itemize}
    \item We introduce \textbf{semantic compression}, a retrieval objective focused on selecting diverse and representative results beyond local proximity.
    \item We formalize semantic compression using submodular optimization and demonstrate how it generalizes standard top-$k$ retrieval.
    \item We propose \textbf{graph-augmented retrieval} by enriching vector spaces with symbolic edges to support multi-hop, context-aware search.
    \item We conduct extensive experiments on both semantic compression and graph-based retrieval, showing that these methods significantly improve semantic diversity while maintaining high relevance.
\end{itemize}

\begin{figure*}[h]
    \centering
    \includegraphics[width=0.95\linewidth]{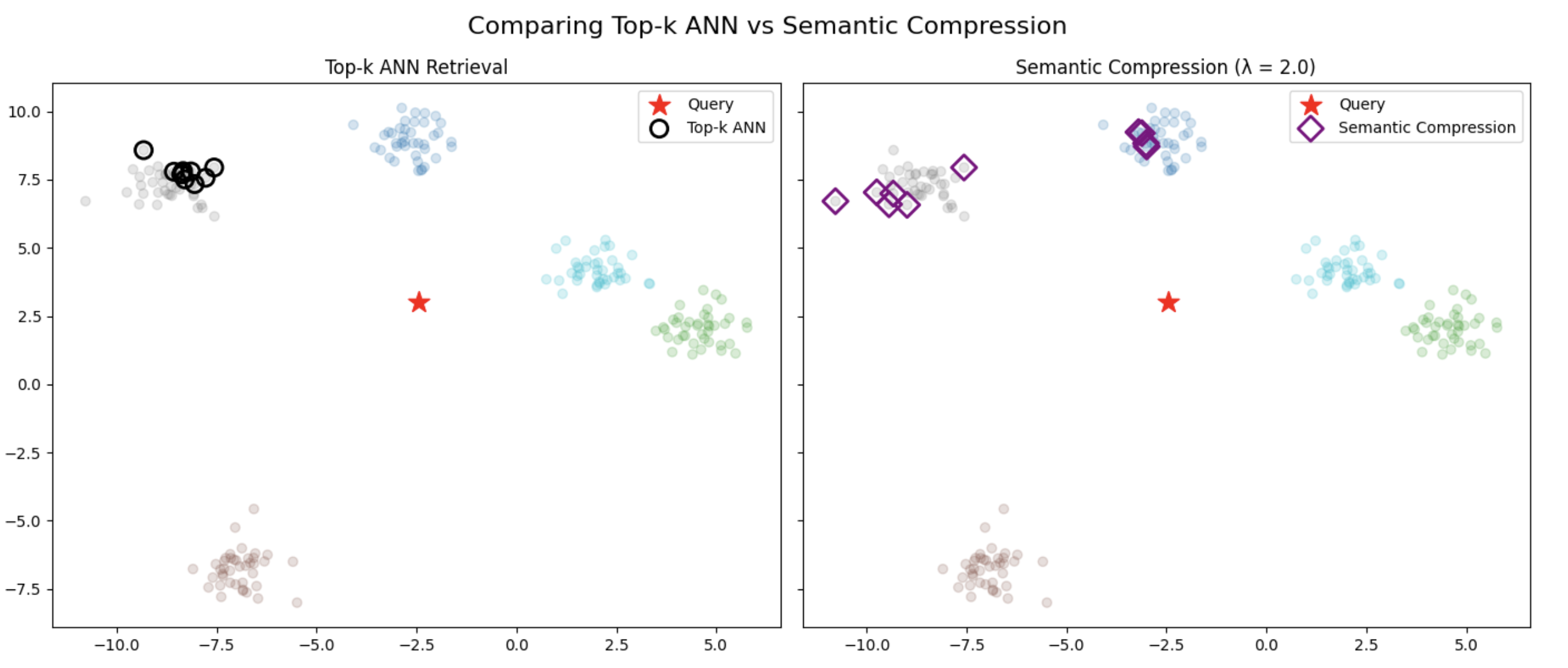}
    \caption{Comparison of Top-$k$ ANN (left) and Semantic Compression (right) in 2D. The query is marked in red. While ANN retrieves points from a single dense region, semantic compression surfaces points across multiple clusters.}
    \label{fig:ann_vs_compression}
\end{figure*}

\section{Related Work}

\textbf{Approximate Nearest Neighbor (ANN) Search}\\
ANN methods form the backbone of modern vector databases, enabling fast retrieval in high-dimensional spaces~\citep{muja2014scalable,johnson2019billion}. Classical approaches include locality-sensitive hashing (LSH)~\citep{indyk1998approximate}, inverted file indexing with product quantization (IVF-PQ)~\citep{jegou2011product}, and graph-based structures such as HNSW~\citep{malkov2018efficient}. While these techniques provide low-latency top-$k$ retrieval, they are primarily designed to minimize geometric distance and do not explicitly account for semantic diversity or coverage.

\textbf{Diversity-Promoting Retrieval}\\
Several works in information retrieval have explored promoting diversity in ranking, notably through maximal marginal relevance (MMR)~\citep{carbonell1998use}, determinantal point processes (DPPs)~\citep{kulesza2012determinantal}, and submodular selection~\citep{krause2014submodular}. However, these approaches are often applied as post-processing steps over retrieved candidates, rather than being integrated into the retrieval mechanism itself. Our work differs by embedding diversity into the retrieval objective and infrastructure directly, motivated by semantic compression.

\textbf{Graph-Augmented Retrieval}\\
Incorporating structure into retrieval systems via graphs has gained interest in recent years. Graph-based ANN methods like HNSW~\citep{malkov2018efficient} implicitly exploit proximity graphs, while knowledge graphs~\citep{cui2019kbqa} and co-occurrence networks~\citep{asai2020learning} have been used to guide multi-hop reasoning. Our proposal builds on this idea by explicitly augmenting vector space with semantic graphs that support multi-hop, context-aware search beyond simple distance metrics.

\textbf{Retrieval for Language Models}\\
Retrieval-augmented generation (RAG)~\citep{lewis2020retrieval} and similar architectures rely heavily on vector search to provide factual grounding to LLMs. However, these systems often retrieve semantically redundant passages, leading to hallucinations or missed evidence~\citep{shi2023replug}. Our method offers a structured retrieval alternative that emphasizes representational coverage, potentially improving downstream performance in LLM pipelines.

\textbf{Comparison to Our Work}\\
While prior research addresses individual components—fast retrieval, diversity, or graph-based augmentation—our work provides a unified framework for semantic compression and graph-augmented retrieval. We contribute theoretical foundations, objective formulations, and architectural implications for next-generation vector search systems.

\section{Semantic Compression}

\subsection{Motivation}
Traditional vector retrieval retrieves the top-\(k\) vectors nearest to a query point \(q\) based on a similarity function (e.g., cosine or dot product)~\citep{johnson2019billion, muja2014scalable}. However, this often results in semantically redundant items—vectors that are close in space but convey overlapping information~\citep{ash2021sampling}. This is especially problematic in tasks where diversity and broad coverage of related concepts are essential, such as open-domain question answering~\citep{asai2020learning} or document summarization~\citep{carbonell1998use, xie2022unified}.

To address this limitation, we introduce the notion of \textit{semantic compression}: the task of selecting a small set of vectors that captures the most semantically informative, diverse, and representative content from the neighborhood of a query.

\subsection{Problem Definition}

Let \( \mathcal{V} = \{v_1, v_2, \dots, v_n\} \subset \mathbb{R}^d \) denote a set of candidate vectors retrieved by an initial ANN pass around a query vector \( q \in \mathbb{R}^d \), and let \( S \subseteq \mathcal{V} \) be a subset of size \(k\) to be returned to the downstream model or user. The goal of semantic compression is to construct a subset \(S\) that is both representative of the semantic content near \(q\) and internally diverse.

We formalize this by the following objective:
\[
\max_{S \subseteq \mathcal{V}, |S| = k} \underbrace{\sum_{v \in \mathcal{V}} \max_{s \in S} \text{sim}(v, s)}_{\text{Coverage}} + \lambda \cdot \underbrace{\sum_{\substack{u, v \in S \\ u \neq v}} (1 - \text{sim}(u, v))}_{\text{Diversity}}
\]
where:
\begin{itemize}
    \item \( \text{sim}(u, v) \) is a similarity measure (e.g., cosine similarity) between vectors \( u \) and \( v \).
    \item The \textbf{coverage term} encourages selection of vectors that "cover" the semantic space of \( \mathcal{V} \) by ensuring each point \(v \in \mathcal{V}\) is similar to at least one item in \(S\).
    \item The \textbf{diversity term} penalizes selection of semantically similar items, promoting representation of different subregions or concepts.
    \item \( \lambda \geq 0 \) balances the trade-off between semantic fidelity and representational spread.
\end{itemize}

\begin{figure*}[h]
    \centering
    \includegraphics[width=0.48\linewidth]{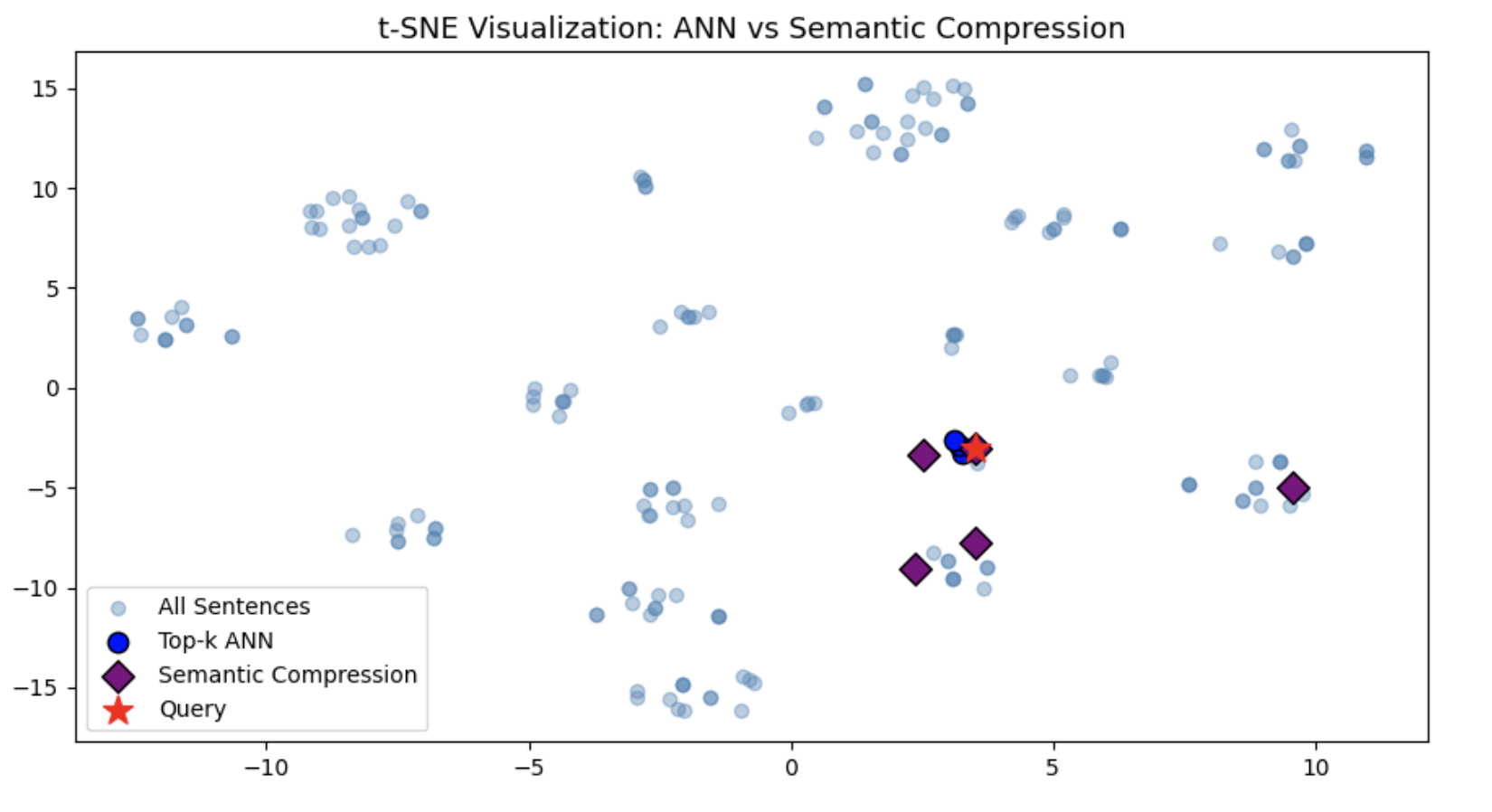}
    \hfill
    \includegraphics[width=0.48\linewidth]{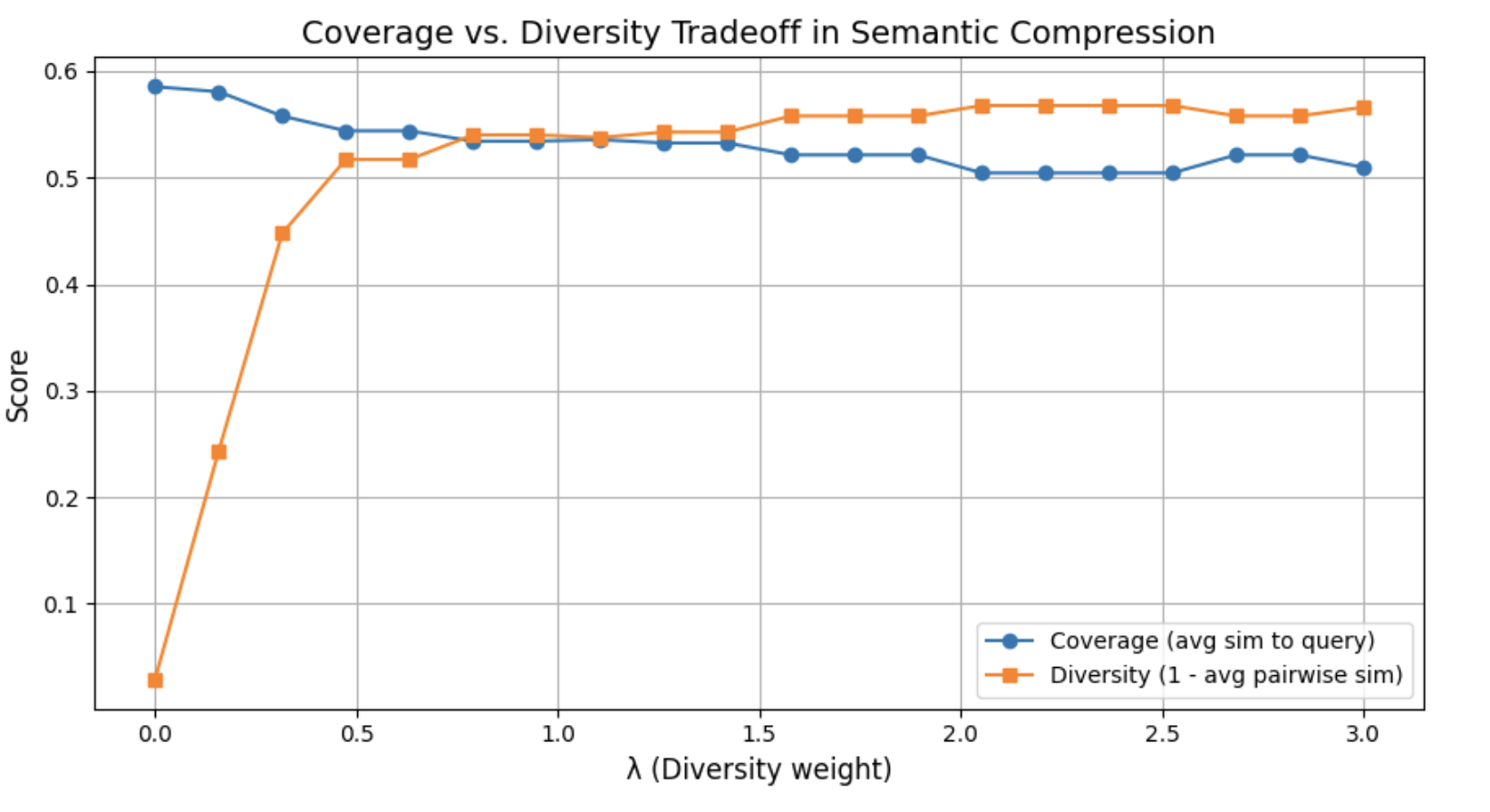}
    \caption{
    \textbf{Left:} t-SNE projection of sentence embeddings, comparing Top-$k$ ANN (blue) with Semantic Compression (purple diamonds) for a single query (red star). While ANN retrieves a tightly clustered set, semantic compression surfaces points from multiple semantic regions.
    \textbf{Right:} Tradeoff between coverage (similarity to query) and diversity (pairwise dissimilarity) as the diversity weight $\lambda$ increases. As $\lambda$ grows, the selection becomes more diverse with only a modest drop in query relevance.
    }
    \label{fig:tsne_tradeoff}
\end{figure*}

\vspace{0.5em}

\subsection{Submodular Formulations and Optimization}

The semantic compression problem can be cast as a submodular maximization task. Submodular functions exhibit diminishing returns, making them amenable to efficient greedy approximations. Let the utility function \( f(S) \) be:

\[
f(S) = \sum_{v \in \mathcal{V}} \max_{s \in S} \text{sim}(v, s) + \lambda \cdot \sum_{\substack{u, v \in S \\ u \neq v}} (1 - \text{sim}(u, v))
\]

The first term behaves like a facility-location function, measuring how well \(S\) represents \(\mathcal{V}\). The second term behaves like a diversity-regularizer based on pairwise distances in the similarity space.

To optimize this objective, a greedy algorithm can be employed:
\begin{itemize}
    \item Initialize \( S \leftarrow \emptyset \).
    \item Iteratively add the element \( v^* \in \mathcal{V} \setminus S \) that maximizes the marginal gain: \( f(S \cup \{v^*\}) - f(S) \).
    \item Repeat until \( |S| = k \).
\end{itemize}

This greedy method achieves a \( (1 - 1/e) \)-approximation guarantee for monotonic submodular functions, and is efficient for real-world deployment.

\subsection{Connection to Retrieval Systems}

This formulation subsumes standard top-$k$ ANN retrieval as a special case. When $\lambda = 0$, the objective reduces to selecting the $k$ vectors with highest similarity to the query, equivalent to nearest neighbor retrieval. As $\lambda$ increases, the selection balances relevance with diversity, promoting coverage across semantically distinct regions. This tradeoff allows the retrieval process to surface results that span multiple subtopics or latent dimensions, rather than concentrating in a single high-density cluster.

\subsection{Implementation of Semantic Compression in Practice}

Semantic compression can be deployed as a lightweight second-stage reranking module following an approximate nearest neighbor (ANN) retrieval step. Below, we describe a concrete implementation pipeline that integrates seamlessly into existing vector retrieval systems.

\paragraph{Step 1: Candidate Generation via ANN.}
Given a query vector \( q \in \mathbb{R}^d \), retrieve an initial candidate pool of size \(N\) using an ANN engine (e.g., FAISS, HNSW):
\[
\mathcal{V} = \{v_1, v_2, \dots, v_N\}, \quad v_i \in \mathbb{R}^d
\]

\paragraph{Step 2: Similarity Matrix Computation.}
Construct:
\begin{itemize}
    \item A query-to-candidate similarity vector:
    \[
    s_q = [\text{sim}(q, v_1), \dots, \text{sim}(q, v_N)] \in \mathbb{R}^N
    \]
    \item A full candidate-to-candidate similarity matrix:
    \[
    S = VV^\top \in \mathbb{R}^{N \times N}
    \]
    where \( V \in \mathbb{R}^{N \times d} \) contains row-stacked normalized vectors. Cosine similarity can be used as the scoring function.

\end{itemize}

\paragraph{Step 3: Subset Selection via Greedy Optimization.}
Using the utility function defined in Section~3.3 (Submodular Formulation), apply the greedy algorithm to select a compressed subset \( S \subset \mathcal{V} \), of size \(k\), that maximizes semantic coverage and diversity.

\begin{itemize}
    \item \textbf{Performance:} For typical values \(N=100\), \(k=10\), greedy selection takes less than 1 ms on modern hardware.
    \item \textbf{Batching:} Matrix operations (\( S = VV^\top \)) and masked indexing enable efficient batching across queries on GPU.
    \item \textbf{Modularity:} This reranking layer is model-agnostic and plug-and-play, requiring no retraining.
    \item \textbf{Extensibility:} Additional relevance scores (e.g., from a graph-based propagation model) can be blended with similarity matrices for hybrid reranking.
\end{itemize}

This implementation strategy makes semantic compression practical for real-time inference settings, enabling improved retrieval without changes to the core ANN infrastructure.

\subsection{Experiment: Comparing ANN and Semantic Compression}

To evaluate the effectiveness of semantic compression, we design a controlled experiment using synthetically generated data. A set of 200 two-dimensional vectors is sampled from five well-separated clusters, simulating distinct semantic regions. A query vector is constructed as the mean of one representative point from each cluster, representing a composite concept that spans multiple topics.

We compute cosine similarity between the query and all candidate vectors and extract the top-50 most similar points to form a candidate retrieval pool. From this pool, we compare two selection strategies for identifying a subset of $k=10$ vectors: (1) standard top-$k$ retrieval based solely on similarity, and (2) semantic compression that balances similarity with diversity across clusters.

Figure~\ref{fig:ann_vs_compression} shows the retrieved vectors under both strategies. While the top-$k$ method tends to concentrate on a single cluster, semantic compression produces a more diverse set spanning all five clusters. This demonstrates the model's ability to capture broader semantic coverage without sacrificing relevance.

To analyze the tradeoff between relevance and diversity in semantic compression, we present two complementary visualizations: a t-SNE projection of the embedding space and a coverage-diversity tradeoff plot.

In Figure~\ref{fig:tsne_tradeoff} (left), Top-$k$ ANN retrieves points densely clustered near the query, often resulting in redundant selections. Semantic compression, on the other hand, produces a more spatially distributed set that captures multiple semantic modes. This leads to improved representational coverage, which is beneficial for downstream tasks such as reranking or retrieval-augmented generation.

Figure~\ref{fig:tsne_tradeoff} (right) illustrates how the method balances similarity and diversity as the diversity weight $\lambda$ increases. When $\lambda = 0$, the method reduces to standard nearest neighbor retrieval. As $\lambda$ increases, diversity rises steadily while coverage remains relatively stable.

These visualizations highlight semantic compression as a tunable, model-agnostic approach that enables flexible control over semantic coverage in retrieval settings.

\section{Graph-Augmented Vector Retrieval}

\subsection{Motivation}

While semantic compression improves retrieval diversity, it is fundamentally limited by the local geometry of the embedding space~\citep{poerner2020evaluating}. Real-world semantic relations—such as synonymy, hierarchy, and multi-hop associations—are often non-metric and cannot be fully captured by vector similarity alone~\citep{nickel2017poincare, asai2020learning, bolukbasi2016all}.

To address this, we propose \textit{graph-augmented vector retrieval}, which overlays a semantic graph on top of the embedding space. The graph encodes latent or symbolic relationships between items, enabling retrieval paths that incorporate global structure, contextual dependencies, and long-range reasoning. This hybrid formulation bridges local similarity with higher-order semantic organization.

\subsection{Graph Construction and Traversal}

We construct a semantic graph \( G = (\mathcal{V}, \mathcal{E}) \), where nodes \( \mathcal{V} = \{v_1, v_2, \ldots, v_n\} \subset \mathbb{R}^d \) correspond to embedded items (e.g., documents or entities), and edges \( \mathcal{E} \subseteq \mathcal{V} \times \mathcal{V} \) encode semantic or structural relationships. This graph augments the embedding space with contextual or symbolic structure.

\paragraph{Edge Formation.} Edges in \( G \) can be defined via:
\begin{itemize}
    \item \textbf{kNN-based similarity:} Connect each node to its \( k \) nearest neighbors under cosine similarity:
    \[
    \text{sim}(v_i, v_j) = \frac{v_i^\top v_j}{\|v_i\| \cdot \|v_j\|}.
    \]
    \item \textbf{External relations:} Use hyperlinks, co-occurrence, or citation edges from metadata or interaction logs.
    \item \textbf{Symbolic graphs:} Incorporate curated relationships from knowledge graphs (e.g., ConceptNet, Wikidata).
\end{itemize}

\paragraph{Graph Traversal.} To propagate relevance from a query-seeded set \( \mathcal{V}_q \subset \mathcal{V} \), we perform semantic expansion via graph-based methods:
\begin{itemize}
    \item \textbf{Random walks:} Sample bounded-length walks from \( \mathcal{V}_q \), accumulating visited nodes.
    \item \textbf{Personalized PageRank:}
    \[
    \mathbf{r} = \alpha \cdot \mathbf{s} + (1 - \alpha) \cdot \mathbf{A}^\top \mathbf{r},
    \]
    where \( \mathbf{A} \) is the normalized adjacency matrix and \( \mathbf{s} \) is the query seed vector.
    \item \textbf{Graph neural networks:} Iteratively update node embeddings via:
    \[
    \mathbf{h}_v^{(l+1)} = \text{AGG} \left( \left\{ \mathbf{h}_u^{(l)} : (u, v) \in \mathcal{E} \right\} \right).
    \]
\end{itemize}

This graph-based augmentation introduces multi-hop semantic reasoning and symbolic structure into the retrieval process, addressing limitations of pure distance-based retrieval in complex domains.

\begin{figure*}[h]
    \centering
    \includegraphics[width=0.42\linewidth]{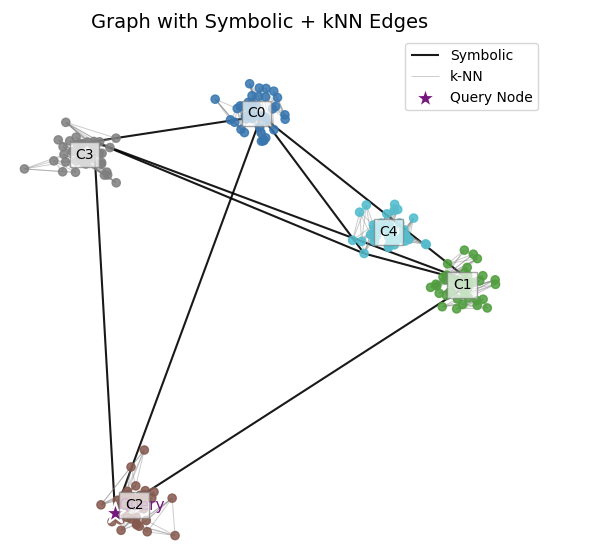}
    \hfill
    \includegraphics[width=0.50\linewidth]{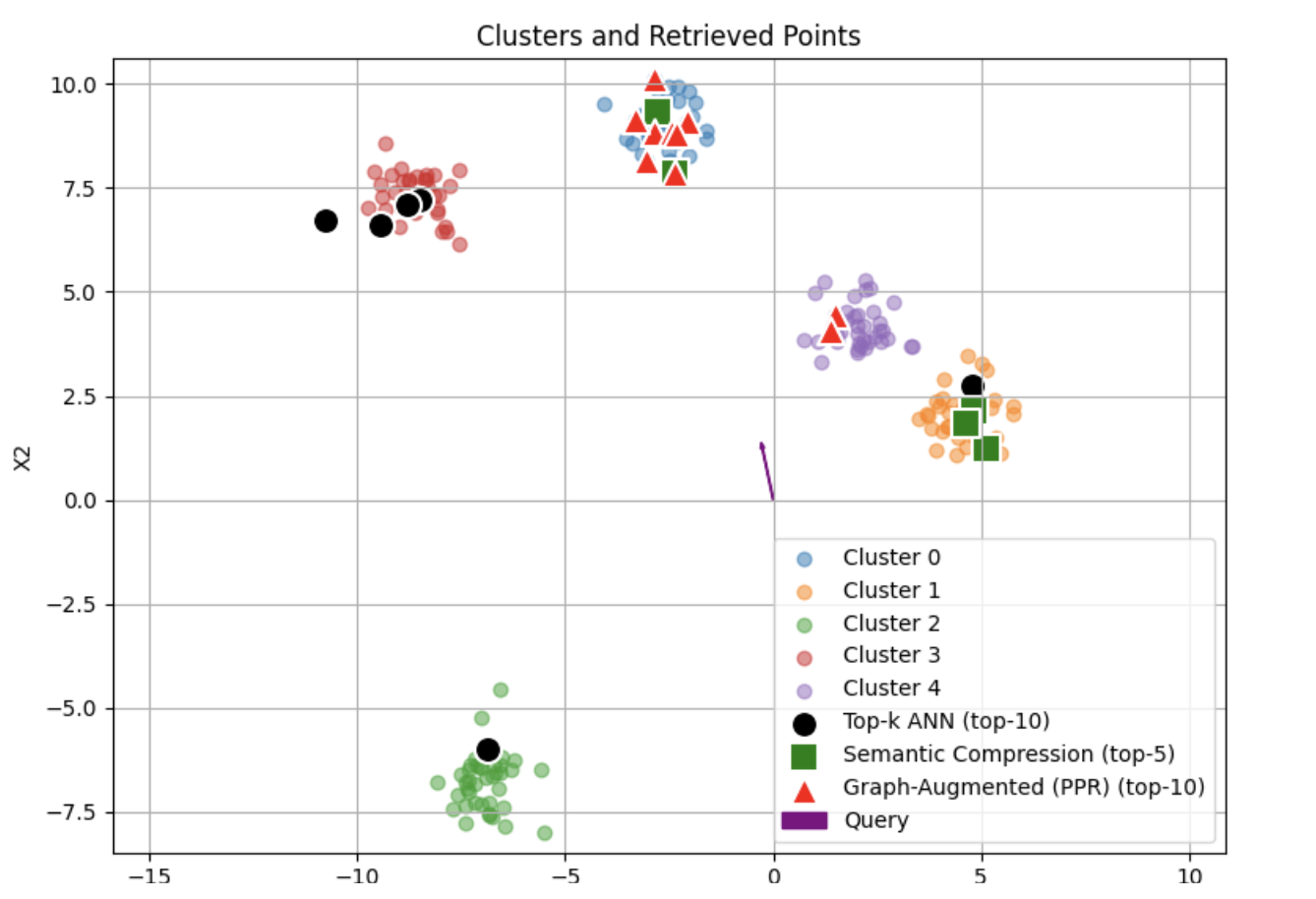}
    \caption{
    \textbf{Left:} Hybrid graph constructed by combining symbolic edges (black) and $k$-nearest neighbor edges (gray). This enriched structure supports semantic exploration through Personalized PageRank.
    \textbf{Right:} Retrieved points for a fixed query (shown in red) across three strategies: Top-$k$ ANN (blue), Semantic Compression (purple diamonds), and Graph-Augmented Retrieval (green). While ANN focuses on local neighborhoods and Semantic Compression samples from diverse clusters, the graph-based method retrieves highly relevant points concentrated in a semantic region.
    }
    \label{fig:graph_cluster_viz}
\end{figure*}

\subsection{Hybrid Scoring and Retrieval}

To bridge the gap between local embedding similarity and global semantic relationships, we introduce a hybrid scoring framework that integrates vector-based relevance with graph-based propagation. Given a semantic graph \( G = (\mathcal{V}, \mathcal{E}) \), and a query embedding \( \mathbf{q} \in \mathbb{R}^d \), we first retrieve a candidate pool \( \mathcal{V}_q \subset \mathcal{V} \) using standard ANN search in the vector space. Our objective is to refine the ranking of nodes in \( \mathcal{V}_q \cup \mathcal{N}(\mathcal{V}_q) \), where \( \mathcal{N}(\cdot) \) denotes neighbors in the graph \( G \), by considering both geometric proximity and graph-induced connectivity.

\paragraph{1. Vector-Based Relevance.}  
The local relevance of a node \( v \in \mathcal{V} \) is defined via cosine similarity between its embedding \( \mathbf{v} \in \mathbb{R}^d \) and the query vector:
\[
S_{\text{vec}}(v, q) = \cos(\mathbf{v}, \mathbf{q}) = \frac{\mathbf{v}^\top \mathbf{q}}{\|\mathbf{v}\| \cdot \|\mathbf{q}\|}.
\]

\paragraph{2. Graph-Based Influence.}  
To capture semantic signals that are not evident from local proximity, we compute a relevance diffusion score \( S_{\text{graph}}(v, q) \) by propagating query affinity through the graph structure. One effective method is \textbf{Personalized PageRank (PPR)}, which computes a stationary distribution over nodes biased towards the query neighborhood:
\[
\mathbf{r} = \alpha \cdot \mathbf{s} + (1 - \alpha) \cdot \mathbf{r} A,
\]
where \( \mathbf{s} \) is a one-hot or soft seed vector centered on \( \mathcal{V}_q \), \( A \) is the normalized adjacency matrix of \( G \), and \( \alpha \in (0,1) \) controls the restart probability. The score \( S_{\text{graph}}(v, q) = \mathbf{r}[v] \) then reflects how reachable \( v \) is from the query region under random walks.

\paragraph{3. Hybrid Scoring Function.}  
We define the final relevance score for each node \( v \in \mathcal{V}_q \cup \mathcal{N}(\mathcal{V}_q) \) as a convex combination of the vector and graph-based components:
\[
R(v \mid q) = (1 - \beta) \cdot S_{\text{vec}}(v, q) + \beta \cdot S_{\text{graph}}(v, q),
\]
where \( \beta \in [0, 1] \) modulates the influence of structural context. This formulation smoothly interpolates between pure ANN retrieval (\( \beta = 0 \)) and purely graph-based exploration (\( \beta = 1 \)).
 
The hybrid approach can be viewed as an instance of \textbf{multi-view retrieval}, where the vector space and graph topology represent two complementary manifolds. The weighted fusion ensures robustness to deficiencies in either view—for example, embedding collapse or graph sparsity.

To ensure scalability:
\begin{itemize}
    \item The graph \( G \) can be restricted to the kNN subgraph over \( \mathcal{V}_q \), significantly reducing memory and compute.
    \item \( S_{\text{graph}} \) can be precomputed for frequent queries or efficiently approximated via truncated random walks.
    \item The scores \( R(v \mid q) \) support batching and GPU acceleration, making them deployable in production retrieval stacks.
\end{itemize}

This hybrid retrieval mechanism significantly enhances semantic generalization, particularly in cases where the query concept has no direct match in the corpus but is indirectly connected via ontological or contextual relationships.

\subsection{Experiment : Graph-Augmented Retrieval with Sparse Symbolic Cluster Connectivity}

We evaluate the three retrieval strategies on a synthetic 2D dataset with clustered structure: (i) Top-$k$ Approximate Nearest Neighbors (ANN), (ii) Semantic Compression using cluster centers, and (iii) Graph-Augmented Retrieval based on Personalized PageRank (PPR) over a symbolic $k$-NN graph.

To assess retrieval quality, we compute two metrics: \textbf{Relevance}, the average cosine similarity between retrieved items and the query, and \textbf{Diversity}, defined as one minus the average pairwise cosine similarity among the retrieved items. Each method retrieves the top-10 results for a fixed query.

\begin{table}[h]
\centering
\begin{tabular}{lcc}
\toprule
\textbf{Method} & \textbf{Relevance} $\uparrow$ & \textbf{Diversity} $\uparrow$ \\
\midrule
Top-$k$ ANN & 0.5174 & 0.8068 \\
Semantic Compression & 0.4798 & 0.5718 \\
Graph-Augmented (PPR) & \textbf{0.9688} & 0.0671 \\
\bottomrule
\end{tabular}
\caption{Comparison of relevance and diversity across retrieval methods.}
\label{tab:results}
\end{table}

Graph-Augmented Retrieval yields the highest relevance by leveraging both symbolic and similarity-based structure. PPR walks tend to remain within semantically consistent regions, which explains the lower diversity. In contrast, Top-$k$ ANN, based on geometric proximity, offers a better balance between relevance and diversity. Semantic Compression provides moderate scores by returning cluster centroids, which capture representative semantics but lose finer granularity.

Figure~\ref{fig:graph_cluster_viz} offers a visual comparison. The left subfigure shows the hybrid symbolic+$k$NN graph used by the PPR-based retrieval, while the right subfigure overlays the retrieved points from each strategy on the original clustered space. We observe distinct retrieval patterns for each method due to their inherent mechanisms. Top-$k$ ANN retrieves items tightly around the query because it directly selects the nearest neighbors in the embedding space based purely on geometric proximity. This approach favors local neighborhoods and can capture fine-grained similarities but may miss semantically diverse points outside the immediate vicinity. In contrast, Semantic Compression samples more broadly across clusters by selecting representative cluster centroids or summaries, thereby promoting diversity by design. However, this broader coverage can reduce relevance since some centroids may lie farther from the query, reflecting a tradeoff that favors diversity over strict query similarity. Finally, Graph-Augmented Retrieval using Personalized PageRank (PPR) focuses on a dense region of semantically aligned points because the diffusion of relevance scores over the hybrid symbolic+$k$NN graph propagates influence through tightly connected nodes. The graph structure encodes both direct semantic relations (symbolic edges) and local similarity (kNN edges), which results in prioritizing nodes strongly connected to the query. This mechanism leads to high relevance but reduces diversity, as the retrieval concentrates within a semantically cohesive neighborhood.

\subsection{Enhancing Graph-Augmented Retrieval via Dense Symbolic Cross-Cluster Connections}

We also explore the impact of significantly increasing the number of symbolic edges by connecting cluster heads to semantically similar nodes across different clusters. Unlike the initial approach, which only linked cluster heads to a limited number of other heads, this enhancement establishes symbolic edges whenever the cosine similarity between a cluster head and any node in a different cluster exceeds a threshold (0.85). This creates a denser and more semantically meaningful symbolic graph, enriching the graph structure with additional cross-cluster connections.
 
\begin{figure*}[h]
    \centering
    \includegraphics[width=0.9\linewidth]{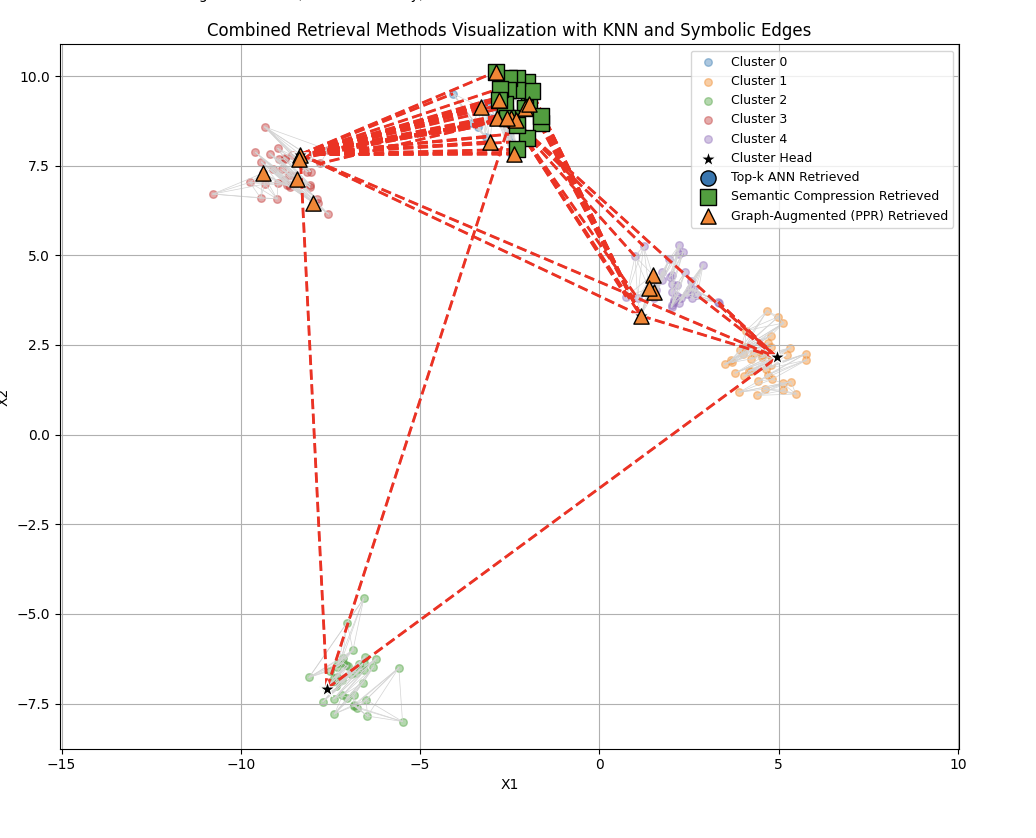}
    \caption{
    Visualization of combined retrieval methods on the dataset with enhanced symbolic edges.
    KNN edges are shown in light gray, while symbolic edges are highlighted as red dashed lines. Cluster heads are marked by black stars.
    Retrieved points from Top-k ANN, Semantic Compression, and Graph-Augmented Retrieval (PPR) are shown with distinct markers and colors.
    }
    \label{fig:combined_retrieval}
\end{figure*}

Figure~\ref{fig:combined_retrieval} illustrates the combined retrieval graph with both k-nearest neighbor (KNN) edges and these expanded symbolic edges. KNN edges are shown in light gray, whereas symbolic edges are highlighted in red dashed lines. Cluster heads are marked as prominent black stars. This visualization clearly shows the enriched symbolic connectivity bridging distant but semantically related points.

\begin{table}[h]
    \centering
    \begin{tabular}{lcc}
        \toprule
        \textbf{Method} & \textbf{Relevance} & \textbf{Diversity} \\
        \midrule
        Top-k ANN & 0.9987 & 0.0013 \\
        Semantic Compression & 0.9987 & 0.0013 \\
        Graph-Augmented(PPR) & 0.9168 & 0.1590 \\
        \bottomrule
    \end{tabular}
    \caption{Relevance and diversity scores for different retrieval methods with enhanced symbolic edges.}
    \label{tab:retrieval_scores}
\end{table}

The results reveal a clear trade-off between relevance and diversity. Both Top-k ANN and Semantic Compression methods achieve very high relevance ($\approx 0.999$) but at the cost of extremely low diversity ($\approx 0.0013$), indicating their retrieved points are highly focused and similar. In contrast, Graph-Augmented Retrieval using Personalized PageRank (PPR) exhibits slightly lower relevance (0.9168) but substantially higher diversity (0.1590). This suggests that incorporating expanded symbolic edges enables the graph to better capture semantic relationships across clusters, leading to richer and more diverse retrieval results.

The expanded symbolic connectivity effectively bridges different clusters by linking semantically related points that are not necessarily nearest neighbors in the original vector space. This enriched graph structure facilitates the PPR algorithm to traverse meaningful semantic pathways beyond local neighborhoods, enhancing the diversity of retrieved results without a significant drop in relevance.

In summary, increasing symbolic edges provides a promising direction to balance relevance and diversity in graph-augmented retrieval systems, surpassing the limitations of purely nearest-neighbor based approaches.

\section{Conclusion}

We presented a new retrieval paradigm that goes beyond traditional top-$k$ nearest neighbor search by prioritizing semantic diversity and representational coverage. Our approach, semantic compression, formalizes retrieval as a submodular optimization problem, enabling selection of compact yet informative results. We further extended this idea with graph-augmented retrieval, integrating symbolic edges into vector space to support multi-hop, context-aware search. Empirical results across diverse retrieval configurations demonstrate that our methods significantly improve semantic diversity without sacrificing relevance. This work lays the foundation for more meaning-aware vector retrieval systems, with applications spanning RAG, question answering, and agent memory retrieval. Future work will explore adaptive graph construction and tighter integration with large language model pipelines.

\section{Impact Statement}

Our work introduces a retrieval framework that prioritizes semantic diversity and contextual relevance, addressing a core limitation of current vector search systems. By formalizing semantic compression and integrating symbolic graph structures, we enable retrieval methods better aligned with the needs of modern language models and multi-step reasoning tasks. This paradigm has broad implications for the design of future retrieval-augmented systems, including factual grounding, multi-hop question answering, and memory-augmented agents. It encourages the research community to rethink retrieval beyond geometric proximity and move toward meaning-centric information access.

\nocite{langley00}

\bibliography{example_paper}
\bibliographystyle{icml2025}

%%%%%%%%%%%%%%%%%%%%%%%%%%%%%%%%%%%%%%%%%%%%%%%%%%%%%%%%%%%%%%%%%%%%%%%%%%%%%%%
%%%%%%%%%%%%%%%%%%%%%%%%%%%%%%%%%%%%%%%%%%%%%%%%%%%%%%%%%%%%%%%%%%%%%%%%%%%%%%%
% APPENDIX
%%%%%%%%%%%%%%%%%%%%%%%%%%%%%%%%%%%%%%%%%%%%%%%%%%%%%%%%%%%%%%%%%%%%%%%%%%%%%%%
%%%%%%%%%%%%%%%%%%%%%%%%%%%%%%%%%%%%%%%%%%%%%%%%%%%%%%%%%%%%%%%%%%%%%%%%%%%%%%%
\newpage
% \appendix
% \onecolumn

% You can have as much text here as you want. The main body must be at most $8$ pages long.
% For the final version, one more page can be added.
% If you want, you can use an appendix like this one.  

% The $\mathtt{\backslash onecolumn}$ command above can be kept in place if you prefer a one-column appendix, or can be removed if you prefer a two-column appendix.  Apart from this possible change, the style (font size, spacing, margins, page numbering, etc.) should be kept the same as the main body.
%%%%%%%%%%%%%%%%%%%%%%%%%%%%%%%%%%%%%%%%%%%%%%%%%%%%%%%%%%%%%%%%%%%%%%%%%%%%%%%
%%%%%%%%%%%%%%%%%%%%%%%%%%%%%%%%%%%%%%%%%%%%%%%%%%%%%%%%%%%%%%%%%%%%%%%%%%%%%%%

\end{document}